\documentclass[11pt]{article}
\usepackage{acl}
\usepackage{times}
\usepackage{latexsym}
\usepackage[T1]{fontenc}
\usepackage[utf8]{inputenc}
\usepackage{microtype}
\usepackage{inconsolata}
\usepackage{graphicx}
\usepackage{booktabs}
\usepackage{multirow}
\usepackage[table]{xcolor}

\title{Lost in Diffusion: Uncovering Hallucination Patterns and Failure Modes in Diffusion Large Language Models}

\author{
 \textbf{Zhengnan Guo\textsuperscript{1,2}},
 \textbf{Fei Tan\textsuperscript{1}$^{*}$}
\\
\\
 \textsuperscript{1}East China Normal University
 \\
 \textsuperscript{2}Zhejiang University of Technology
 \\
    {\normalsize \texttt{\href{mailto:zhengnan.guo@zjut.edu.cn}zhengnan.guo@zjut.edu.cn}, \texttt{\href{mailto:ftan@mail.ecnu.edu.cn}ftan@mail.ecnu.edu.cn}}
}

\begin{document}
\maketitle

\renewcommand{\thefootnote}{}
\footnotetext{$^{*}$ Corresponding author}
\renewcommand{\thefootnote}{\arabic{footnote}}

\begin{abstract}
While Diffusion Large Language Models (dLLMs) have emerged as a promising non-autoregressive paradigm comparable to auto-regressive (AR) models, their faithfulness, specifically regarding hallucination, remains largely underexplored. To bridge this gap, we present the first controlled comparative study to evaluate hallucination patterns in dLLMs. Our results demonstrate that current dLLMs exhibit a higher propensity for hallucination than AR counterparts controlled for architecture, scale, and pre-training weights. Furthermore, an analysis of inference-time compute reveals divergent dynamics: while \emph{quasi-autoregressive} generation suffers from early saturation, \emph{non-sequential} decoding unlocks potential for continuous refinement. Finally, we identify distinct failure modes unique to the diffusion process, including \emph{premature termination}, \emph{incomplete denoising}, and \emph{context intrusion}. Our findings underscore that although dLLMs have narrowed the performance gap on general tasks, their distinct hallucination mechanisms pose a critical challenge to model reliability. Our code is available at \href{https://github.com/ZeroLoss-Lab/Lost-in-Diffusion}{https://github.com/ZeroLoss-Lab/Lost-in-Diffusion}
\end{abstract}
\section{Introduction}

While Auto-regressive Large Language Models (AR-LLMs) have dominated text generation in recent years \cite{openai2022chatgpt,grattafiori2024llama3herdmodels,deepseekai2025deepseekv3technicalreport}, a growing body of research is exploring Diffusion Large Language Models (dLLMs) as a potential non-autoregressive generation paradigm \cite{deepmind2024gemini_diffusion,inceptionlabs2025mercury,song2025seeddiffusionlargescalediffusion}. Diverging from the sequential token generation of AR models, dLLMs generate tokens in parallel via iterative denoising, offering inherent advantages in global planning and bidirectional visibility \cite{li2022diffusionlm,gong2023diffuseq}. Recent open-source dLLMs, such as LLaDA series \cite{nie2025large, bie2025llada20scalingdiffusionlanguage}, Dream \cite{ye2025dream}, SDAR \cite{cheng2025sdarsynergisticdiffusionautoregressionparadigm} and Fast-dLLM v2 \cite{wu2025fastdllmv2efficientblockdiffusion}, have achieved performance parity with leading AR-LLMs at comparable scales, signaling that dLLMs are approaching practical utility.

Despite these advancements, the trustworthiness of dLLMs, specifically regarding hallucination, remains an open question. Hallucination, defined as inconsistency between model output and source content (i.e., input context or training data) \cite{Ji2023Surcey, bang-etal-2025-hallulens}, is well-documented in AR-LLMs \cite{Huang2023ASO,Kalai2025WhyLM}. These inconsistencies typically stem from sequential error propagation and unidirectional attention mechanisms, phenomena respectively known as the snowballing effect~\cite{zhang2024how} and the reversal curse~\cite{berglund2024the}. Theoretically, the global context planning and bidirectional visibility of dLLMs offer a mitigation by facilitating retrospective refinement. Conversely, the stochastic nature of the diffusion process introduces intrinsic noise, potentially exacerbating decoding randomness, a factor also widely recognized as a root cause of hallucination~\cite{leefactuality}.  As dLLMs move towards widespread adoption, a critical question emerges: \emph{Does the diffusion mechanism mitigate or exacerbate hallucination?}

To address this, we conduct the first systematic benchmarking of dLLMs against their AR counterparts regarding hallucination. Our contributions are threefold:

\begin{itemize}
    \item Benchmarking the Mechanism: We provide the first controlled pairwise comparison, revealing that current dLLMs exhibit a significantly higher propensity for hallucination than AR-LLMs under comparable settings.
    \item Dynamics of Inference Compute: We find that the efficacy of denoising is contingent upon the decoding strategy. Specifically, while \emph{quasi-autoregressive} generation suffers from early saturation, \emph{non-sequential} strategies unlock the potential for continuous refinement via increased inference steps.
    \item Diffusion Failure Modes: We identify and analyze failure modes unique to the diffusion process, including \emph{premature termination}, \emph{incomplete denoising}, and \emph{context intrusion}, offering insights into their underlying mechanisms.
\end{itemize}

\section{Related Work}

\subsection{Diffusion Large Language Models}
Early text diffusion attempts \cite{li2022diffusionlm,zhang2023planner,Lin2023Pre-trainning,gulrajani2023likelihoodbased} struggled with the discrete nature of language. However, recent advancements in masked diffusion \cite{Lou2023DiscreteDM,shi2024simplified,sahoo2024simple} have enabled significant scaling. Representative open-source models, including the LLaDA series \cite{nie2025large, bie2025llada20scalingdiffusionlanguage}, Dream \cite{ye2025dream}, SDAR \cite{cheng2025sdarsynergisticdiffusionautoregressionparadigm}, and Fast-dLLM v2 \cite{wu2025fastdllmv2efficientblockdiffusion}, have achieved performance parity with AR-LLMs on general benchmarks. Despite this progress, current research predominantly prioritizes architectural optimization and inference acceleration. While \citet{chang2025tracedet} recently initiated the study of hallucination detection in dLLMs, a systematic comparative benchmarking of their hallucination patterns against AR baselines remains absent.

\subsection{Hallucination in LLMs}
Hallucination constitutes a primary obstacle to the reliable deployment of LLMs \cite{Ji2023Surcey,Huang2023ASO}. In AR-LLMs, this phenomenon is often traced to the sequential decoding mechanism. Specifically, the snowballing effect \cite{zhang2024how} illustrates how early sequence errors propagate and amplify, while the reversal curse \cite{berglund2024the} exposes the structural inability of auto-regressive objectives to handle bidirectional knowledge retrieval. Additionally, the stochastic nature of sampling correlates with decoding randomness and factual inconsistencies \cite{leefactuality}. While these patterns are well-charted in AR models, dLLMs exhibit distinct failure modes, such as interleaving hallucinations \cite{chang2025tracedet}. Motivated by these distinctions, we present the first controlled study to quantify hallucinations in dLLMs in comparison to AR baselines.

\section{Methodology}

To assess the hallucination patterns of dLLMs, we propose a comparative framework designed to minimize confounding factors such as model scale and training data distribution. While complete isolation of the generation mechanism is inherently challenging, our methodology represents a best-effort approach to disentangling it from these variables.

\subsection{Pairwise Comparison Framework}
\label{sec:pair}

Direct comparisons between dLLMs and arbitrary AR-LLMs are often skewed by discrepancies in pre-training corpora and model capacities, as these discrepancy can substantially affect model performance even under identical task settings~\cite{wang-etal-2024-scaling, lu-etal-2023-makes}. To mitigate this, we devise a pairwise comparison strategy comprising two distinct control groups, which aims to maximize comparability by aligning parametric knowledge and scale to the extent possible:

\begin{itemize}
    \item \textbf{Group I: Architectural Alignment.} We benchmark LLaDA-8B \cite{nie2025large} against LLaMA-3-8B \cite{grattafiori2024llama3herdmodels}. Both models share similar backbone architectures and parameter scales, and exhibit comparable performance on general benchmarks, allowing us to isolate differences induced specifically by the diffusion modeling approach.
    \item \textbf{Group II: Parametric Alignment.} We pair Dream-7B \cite{ye2025dream} with Qwen2.5-7B \cite{qwen2025qwen25technicalreport}. Crucially, as Dream is initialized directly from Qwen weights, this pairing offers the control over parametric knowledge. Any divergence in hallucination rates can thus be primarily attributed to the diffusion generation process rather than knowledge disparity.
\end{itemize}

We prioritize pre-trained checkpoints for our primary analysis to isolate the generation paradigm from the noise introduced by post-training alignment, although instruction-tuned variants are discussed in Appendix \ref{sec:appendix}. Note that only the instruction-tuned checkpoints are publicly released for LLaDA2.0 and Fast-dLLM v2.

\subsection{Canonical Diffusion Inference}

Unlike the sequential dependence of AR-LLMs ($x_t \sim p(x_t | x_{<t})$), dLLMs generate the entire sequence in parallel via iterative denoising. To fully characterize this mechanism, we adopt a \emph{canonical diffusion} setting, bypassing semi-autoregressive or block-based acceleration methods \cite{nie2025large,wu2025fastdllmv2efficientblockdiffusion}.

For a target sequence of length $L$, we set the number of denoising steps $T$ equal to the sequence length ($T=L$). This configuration maximizes the model's capacity for iterative refinement. To ensure reproducibility, we set the temperature to zero. We employ standard decoding strategies: high-confidence decoding for LLaDA and minimum entropy decoding for Dream.

\subsection{Hallucination Assessment Protocol}

We adapt the HalluLens \cite{bang-etal-2025-hallulens} benchmark to evaluate \emph{Extrinsic Hallucination} \cite{Ji2023Surcey}, spanning three tasks: \emph{PreciseWikiQA}, \emph{LongWiki}, and \emph{NonExistentRefusal}. Please refer to Appendix \ref{sec:settings} for detailed configurations.

To strictly evaluate the generation mechanism, we exclude \emph{Intrinsic Hallucination} tasks (e.g., summarization). These tasks heavily rely on instruction-following capabilities typically acquired during supervised fine-tuning, which would introduce confounding factors unrelated to the pre-training generation objective. Since our benchmark relies on an automatic LLM-based evaluator, we additionally perform human annotation on a stratified subset to validate its reliability. Detailed information can be found in Appendix~\ref{validation}.
\section{Experiments and Analysis}
\label{sec:main_results}

We systematically evaluate dLLMs against their AR counterparts across three critical dimensions: precise knowledge recall, long-form factual consistency, and knowledge boundary detection. Detailed experiments settings can be found in Appendix \ref{sec:settings}. The comparative results are presented in Table \ref{tab:main_results}.

\subsection{Main Results}

\begin{table*}[t]
\centering
\small
\resizebox{\textwidth}{!}{
\begin{tabular}{l|ccc|ccc|c}
\toprule
\multirow{2}{*}{\textbf{Model}} & \multicolumn{3}{c|}{\textbf{Precise WikiQA}} & \multicolumn{3}{c|}{\textbf{LongWiki}} & \textbf{NonexistentRefusal} \\
 & \textbf{FRR} $\downarrow$ & \textbf{HR} $\downarrow$ & \textbf{CR} $\uparrow$ & \textbf{Prec.} $\uparrow$ & \textbf{Rec.@32} $\uparrow$ & \textbf{F1@32} $\uparrow$ & \textbf{FA} $\downarrow$ \\
\midrule
LLaMA-3-8B & 28.72 & \textbf{85.94} & \textbf{10.30} & \textbf{0.408} & 0.293 & \textbf{0.306} & \textbf{73.35} \\
LLaDA-8B & \textbf{21.40} & 95.13 & 3.92 & 0.271 & \textbf{0.306} & 0.272 & 87.10 \\
\midrule
Qwen2.5-7B & \textbf{18.12} & \textbf{89.06} & \textbf{9.06} & \textbf{0.376} & \textbf{0.441} & \textbf{0.387} & \textbf{94.05} \\
Dream-7B & 26.96 & 92.54 & 6.04 & 0.345 & 0.400 & 0.340 & 98.50 \\
\bottomrule
\end{tabular}
}

\caption{Performance comparison across three extrinsic hallucination tasks. Metrics include False Refusal Rate (FRR), Hallucination Rate (HR), and Correct Rate (CR) for \emph{PreciseWikiQA}; Precision, Recall@32, and F1@32 for \emph{LongWiki}; and False Acceptance Rate (FA) for \emph{NonexistentRefusal}. \textbf{Bold} indicates the better performance within each control group.}
\label{tab:main_results}
\end{table*}

Basically, dLLMs consistently underperform their AR counterparts across all three tasks. In \emph{PreciseWikiQA}, dLLMs struggle to anchor generation to facts, with Dream-7B exhibiting a higher Hallucination Rate (92.54\%) than Qwen2.5-7B (89.06\%) and LLaDA-8B showing a significantly lower Correct Rate (3.92\%) compared to LLaMA-3-8B (10.30\%). In \emph{LongWiki}, while LLaDA-8B achieves competitive Recall, both dLLMs suffer from lower Precision, resulting in inferior overall F1@32 scores. In \emph{NonExistentRefusal}, dLLMs reveal a severe inability to refuse invalid queries; notably, Dream-7B fails to recognize non-existent entities in 98.50\% of cases, lagging behind the AR baseline.

This performance degradation can be partially attributed to the absence of a mature re-masking strategy. While dLLMs theoretically offer global planning, current architectures lack the mechanism to retrospectively correct tokens once they are denoised, even if subsequent context renders them implausible \cite{kang2026parallelbench}. Consequently, the snowballing effect is not eliminated but transformed: early denoising errors become entrenched, corrupting the trajectory of the entire generation process. Furthermore, the reversal curse is not fully mitigated; although bidirectional attention enables the detection of inconsistent tokens, the model remains unable to rectify them. Compounded by the immaturity of current training techniques tailored for dLLMs, these structural deficiencies ultimately exacerbate the issue of hallucinations. More recently, a number of  approaches \cite{wang2025remasking, zhang2026correctivediffusionlanguagemodels, huang2026dont, bie2026llada21speedingtextdiffusion} have been proposed to mitigate token-level errors. Evaluating these methods within our framework is an important direction that we leave to future work.

\subsection{Dynamics of Inference Compute}
\label{sec:step_analysis}

A theoretical advantage of dLLMs is the ability to trade compute for quality via iterative refinement. To investigate whether increased inference compute mitigates hallucination, we evaluate LLaDA-8B and Dream-7B on the \textit{LongWiki} task across exponentially increasing denoising steps $T \in \{128, 256, 512, 1024\}$.

\begin{figure}[t]
    \centering
    \includegraphics[width=\linewidth]{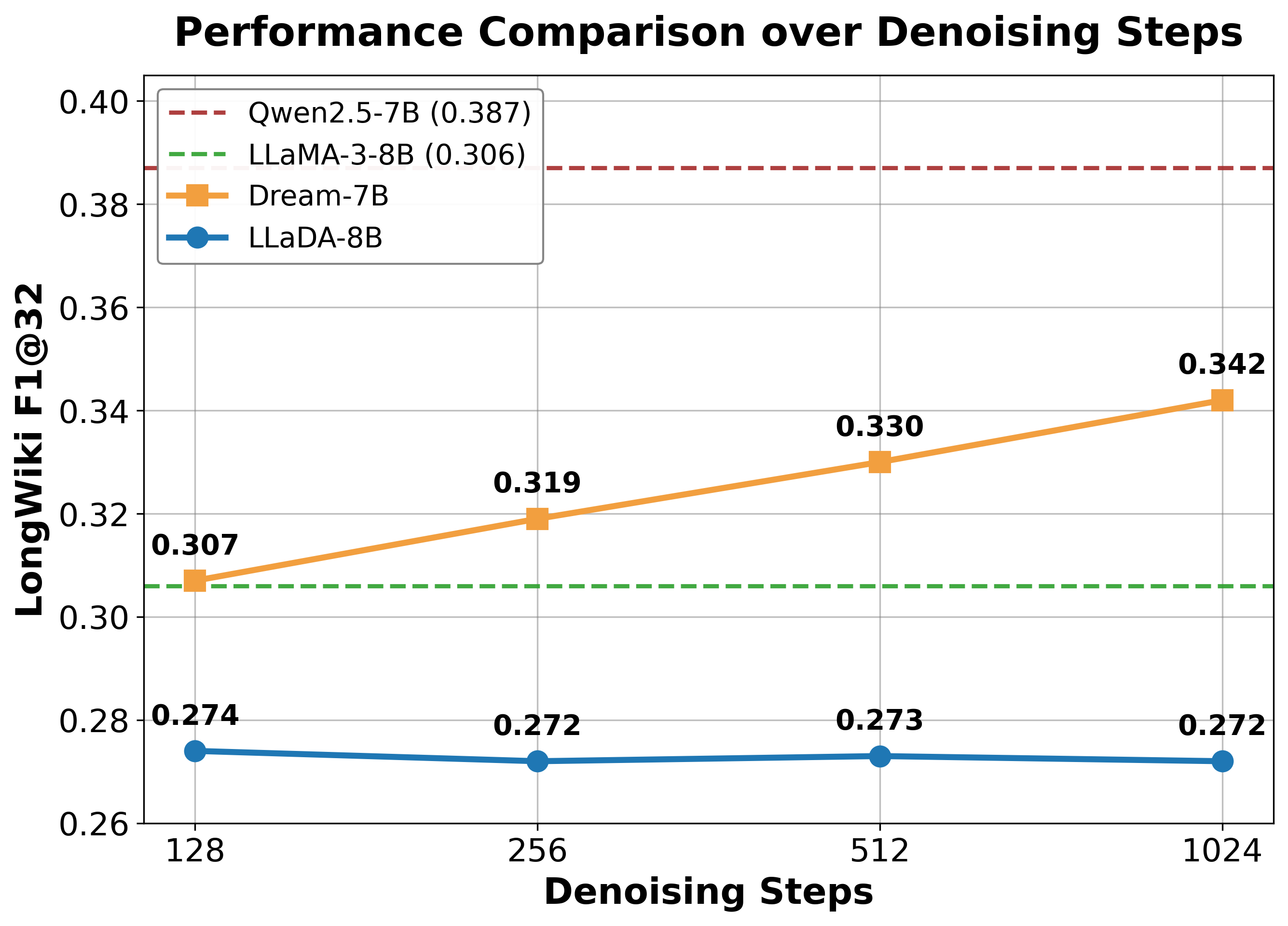}
    \caption{F1@32 trends across exponentially increasing denoising steps. While LLaDA saturates early, Dream benefits from increased inference compute.}
    \label{fig:step_paradox}
\end{figure}

As shown in Figure \ref{fig:step_paradox}, the two architectures exhibit starkly divergent behaviors. LLaDA-8B displays early saturation, with performance metrics remaining statistically stagnant across all steps (F1@32 $\approx$ 0.27). We attribute this to its \emph{quasi-autoregressive} generation order. Despite theoretically possessing bidirectional visibility, LLaDA-8B is constrained by a linear noise scheduler and high-confidence decoding that enforces a predominant left-to-right generation flow (see Appendix \ref{sec:generation_1}). Consequently, it inherits the structural rigidity of AR models while incurring additional diffusion noise. Increasing steps merely reduces prediction granularity without fundamentally altering sequential dependencies, explaining the negligible marginal utility of compute scaling.

In contrast, Dream-7B demonstrates positive scaling dynamics, exhibiting monotonic improvements as $T$ increases. This is driven by its minimum entropy decoding, which enables genuine \emph{non-sequential} refinement. Dream-7B breaks the left-to-right paradigm, more effectively exploiting bidirectional self-attention (see Appendix \ref{sec:generation_2}). This allows for a significant reduction in hallucination rates as compute budget increases. While this flexibility unlocks the potential for inference-time scaling, it also introduces distinct stability risks. Our analysis reveals that departing from sequential constraints is closely associated with a higher incidence of \textit{Context Intrusion}, shifting the distribution of failure modes compared to LLaDA. We categorize these unique failure modes below.

\section{Diffusion Failure Modes}
\label{sec:failure_modes}

Unlike the sequential drift observed in AR-LLMs, dLLMs exhibit unique failure modes stemming from their non-autoregressive nature. Through manual inspection, we identify three distinct patterns (see Appendix \ref{sec:Failure_Modes_Examples} for case studies):

\noindent\textbf{Premature Termination:} 
The model predicts the End-of-Sequence (EOS) token or inserts rigid separators abruptly. This typically occurs when disjoint segments (e.g., the prefix and suffix) decoded independently fail to align syntactically. Lacking a mechanism to bridge this coherence gap, the model is forced to terminate or insert broken structures to resolve the conflict.

\noindent\textbf{Incomplete Denoising:} 
In scenarios involving rare entities, the model may anchor on nonsensical tokens in the later part of the sequence. As bidirectional attention attempts to rationalize the connection between the prompt and these chaotic anchors, the entire generation collapses. This results in residual traces or hollow symbols that imitate the structure of language while lacking any substantive content.

\noindent\textbf{Context Intrusion:} 
We observe instances where the model abruptly pivots to unrelated topics, such as mathematical reasoning or coding tutorials. This phenomenon is triggered when the model occasionally denoises a high-frequency token (e.g., a number or code keyword) in a future position. The bidirectional attention then forces the generation to construct a logical path to this spurious anchor, effectively hijacking the original query trajectory.

We further quantify the prevalence of diffusion-specific failure modes through human annotation. We annotate 200 hallucinated outputs from dLLMs and assign each example failure mode label among \textit{Premature Termination} (PT), \textit{Incomplete Denoising} (ID), \textit{Context Intrusion} (CI). As shown in Table~\ref{tab:failure_modes_q}, these failure modes occur at non-trivial frequencies in both dLLMs, with \textit{Incomplete Denoising} more common in LLaDA-8B and \textit{Context Intrusion} more frequent in Dream-7B.

These failure modes expose the inherent cost of the unconstrained generation paradigm. Unlike AR models that only ensure continuity with the past, dLLMs must satisfy global consistency across the entire sequence. When early-decoded tokens conflict, the current absence of iterative editing mechanisms, such as insertion, deletion, or re-masking, leaves the model with no recourse but to force a broken interpretation. This suggests that future dLLMs require dynamic sequence editing capabilities akin to human writing to fully realize the potential of non-autoregressive generation.

\begin{table}[t]
\centering
\begin{tabular}{l c c c}
\toprule
Model & PT & ID & CI \\
\midrule
LLaDA-8B & 18.0 & 60.0 & 38.0\\
Dream-7B & 13.0 & 44.0 & 58.0\\
\bottomrule
\end{tabular}
\caption{Frequency (\%) of human-annotated failure modes on broken outputs from diffusion language models. Percentages do not sum to 100 because some examples exhibit multiple failure modes.}
\label{tab:failure_modes_q}
\end{table}
\section{Conclusion}

We present a controlled evaluation revealing that current dLLMs are generally more prone to hallucination than auto-regressive baselines. Our analysis of inference dynamics reveals a dichotomy: while \emph{quasi-autoregressive} dLLMs suffer from early saturation, \emph{non-sequential} decoding unlocks the potential for continuous refinement but introduces stability risks. Coupled with unique failure modes like \emph{premature termination}, these findings underscore that achieving reliable non-autoregressive generation requires more than simple denoising.

\section*{Limitations}

While our study offers a foundational benchmarking of dLLM hallucinations, we acknowledge several limitations.

\noindent \textbf{Limits of Confounders Isolation} While our comparative framework strives to isolate the impact of the generation paradigm to the maximum extent possible, achieving a strictly controlled environment is theoretically elusive. Even the most lightweight adaptation required to enable diffusion capabilities necessitates weight updates, which inevitably shift the model's underlying parametric knowledge. Consequently, such shifts may introduce subtle confounders in hallucination assessment. Our approach thus represents a best-effort approximation within these inherent constraints.

\noindent \textbf{Scope of Inference Mechanism} In this work, we strictly adopt a \emph{canonical diffusion} setting to characterize the native generation behavior. Consequently, our findings may not fully generalize to accelerated inference methods (e.g., semi-autoregressive or block-based decoding) where auto-regressive guidance is reintroduced to stabilize generation. Furthermore, given that these models were not explicitly trained to perform denoising over such extensive block lengths ($1,024$), our observed hallucination levels may not faithfully reflect those encountered in real-world deployments. Accordingly, our findings may not directly transfer to alternative inference-time control or verification schemes, which have been shown to reshape generation trade-offs in other decoding paradigms~\cite{DBLP:conf/icml/LeviathanKM23, ding-etal-2025-consultant, yan-etal-2025-decoding}.

\noindent \textbf{Exclusion of Alignment Tuning} As mentioned in Section \ref{sec:pair}, we prioritize pre-trained models to isolate the generative paradigm from post-training noise. As noted in our analysis in Appendix \ref{sec:appendix}, instruction-tuned dLLMs currently exhibit high variance in refusal behaviors (e.g., the over-refusal observed in Dream-7B), which confounds the assessment of intrinsic hallucination propensities. For this reason, we excluded instruction-tuned models from our primary analysis; by the same logic, we also omitted \emph{Intrinsic Hallucination} tasks, as they rely heavily on instruction-following capabilities acquired during alignment. However, we acknowledge that a more in-depth evaluation of the impact of alignment tuning on hallucination levels is certainly feasible, and we leave this extensive investigation to future work.

\section*{Acknowledgments}

This work is supported by the East China Normal University “Artificial Intelligence” Seed Grant Program (40500-20101-222438) and the East China Normal University “Discipline Advancement Program” (40600-515100-25001/002/015).

\bibliography{custom}

\appendix
\section{Experimental Settings}
\label{sec:settings}

In this section, we detail the experimental setup for evaluating the hallucination patterns of dLLMs versus AR-LLMs. Our protocol adheres to the HalluLens benchmark \cite{bang-etal-2025-hallulens}, with strategic adaptations to the evaluator backend to ensure state-of-the-art judgment accuracy.

\subsection{Model Inference and Baselines}

To ensure a rigorous comparison between these distinct paradigms, we standardize inference parameters across all models.

\begin{itemize}
    \item \textbf{Inference Parameters:} We set the temperature to $0$ and top-$p$ to $1.0$ for all experiments. This eliminates sampling randomness, ensuring strict reproducibility.
    \item \textbf{Diffusion Setting:} For dLLMs, we employ a \emph{canonical diffusion} setting. The number of denoising steps $T$ is set equal to the sequence length $L$ (i.e., $T=L$). This configuration grants the model maximum theoretical capacity for iterative refinement during the noise-to-data transition.
    \item \textbf{Auto-regressive Baseline:} For AR-LLMs, we utilize standard greedy decoding.
\end{itemize}

\subsection{Automated Evaluation Engine}

While the original HalluLens framework utilizes Llama-3.1-70B/405B-Instruct \cite{grattafiori2024llama3herdmodels} for data generation and judgment, we standardize our pipeline using \textbf{Gemini-2.5-Flash} \cite{comanici2025gemini25pushingfrontier}. We selected Gemini-2.5-Flash for its superior reasoning capabilities and cost-efficiency in processing long-context verification tasks. This model serves three distinct roles:

\begin{itemize}
    \item \textbf{Question Generator:} Constructing dynamic, knowledge-seeking questions derived from reference documents.
    \item \textbf{Claim Extractor:} Decomposing LLM-generated sentences into atomic, independently verifiable claims.
    \item \textbf{LLM-as-a-Judge:} Evaluating the correctness of responses and detecting refusal behaviors.
\end{itemize}

\subsection{Task-Specific Configurations}

\paragraph{PreciseWikiQA.}
This task evaluates the model's ability to recall precise entities from its parametric memory. We limit the output length to 256 tokens. The \emph{GoodWiki} \cite{GoodWiki} dataset serves as the knowledge source, with difficulty controlled by binning Wikipedia pages based on harmonic centrality to ensure a balanced distribution between head and long-tail knowledge. To prevent data contamination, we follow the HalluLens protocol to dynamically generate 5,000 questions. Specifically, we employ Gemini-2.5-Flash to: (1) select a Wikipedia section and generate a specific, fact-seeking question; (2) verify the answerability of the question based solely on the reference text; and (3) \textbf{classify} responses into three categories: \emph{Correct}, \emph{Incorrect}, or \emph{Refusal}. We report the Hallucination Rate (HR) specifically for non-refused answers, alongside the False Refusal Rate (FRR).

\paragraph{LongWiki.} 
This task assesses the factual consistency of long-form generation (1,024 tokens). Analogous to PreciseWikiQA, Gemini-2.5-Flash generates 250 open-ended prompts based on selected Wikipedia pages. We adopt the FactScore-style \cite{min-etal-2023-factscore, song-etal-2024-veriscore, wei2024longform} evaluation pipeline adapted for HalluLens, replacing the backbone models with Gemini-2.5-Flash. The process involves: (1) Gemini-2.5-Flash decomposes the generated text into atomic claims; (2) we retrieve the top-5 most relevant passages from the source Wikipedia page using dense retrieval; and (3) Gemini-2.5-Flash verifies each claim against the retrieved evidence, labeling it as \emph{Supported} or \emph{Not Supported}. Following \citet{bang-etal-2025-hallulens}, we report not only \emph{Precision} \cite{min-etal-2023-factscore} but also \emph{Recall@K} \cite{wei2024longform} and \emph{F1@K} (with $K=32$). While Precision measures the accuracy of generated claims, relying on it alone allows models to ``game'' the metric by generating overly short responses. Recall@K is thus critical to ensure informativeness, while F1@32 provides a holistic view of hallucination.

\paragraph{NonExistentRefusal.}
This task evaluates the model's ability to recognize knowledge boundaries by querying about fabricated entities (output length: 256 tokens). We first synthesize 2,000 non-existent names by recombining existing taxonomic names from the ITIS database \cite{itis_database} and pharmaceutical registries \cite{tiwari2024medicines}, verifying them against the database to ensure non-existence. Subsequently, models are prompted to describe these entities. Finally, Gemini-2.5-Flash acts as the judge to determine if the model treats the entity as real (i.e., \emph{Hallucination/False Acceptance}) or correctly identifies it as non-existent (i.e., \emph{Successful Refusal}). We report the False Acceptance Rate (FA), where a lower score indicates superior refusal capabilities.

\section{Human Validation of Automatic Evaluation}
\label{validation}

\begin{table*}[htbp]
\centering
\resizebox{\textwidth}{!}{\begin{tabular}{l l l c c}
\toprule
Task & Sample size & Metric validated & Human--Auto Agreement & $\kappa$ \\
\midrule
PreciseWikiQA & 100 & Correct / Incorrect / Refusal & 88.0 & 0.74 \\
LongWiki & 300 claims & Supported / Not Supported & 84.3 & 0.68 \\
NonExistentRefusal & 100 & Accept / Refuse & 91.0 & 0.81 \\
Overall & --- & --- & 87.8 & 0.74 \\
\bottomrule
\end{tabular}}
\caption{Human validation of automatic judgments across the three evaluation tasks.}
\label{tab:human_validation}
\end{table*}

We sample 100 examples per task, balancing across models and outcome types. Three annotators with NLP background independently label each example using written guidelines and the corresponding reference evidence. For PreciseWikiQA and NonExistentRefusal, responses are labeled as correct/factual, hallucinated, or refusal. For LongWiki, annotators verify a sampled subset of atomic claims against the retrieved evidence and label each claim as supported or unsupported. We aggregate labels by majority vote and report both inter-annotator agreement and human–automatic agreement. Table~\ref{tab:human_validation} shows the results.

\section{Hallucination in Instruct-tuned dLLMs}
\label{sec:appendix}

\begin{table*}[htbp]
\centering
\small
\resizebox{\textwidth}{!}{
\begin{tabular}{l|ccc|cccc|c}
\toprule
\multirow{2}{*}{\textbf{Model}} & \multicolumn{3}{c|}{\textbf{Precise WikiQA}} & \multicolumn{4}{c|}{\textbf{LongWiki}} & \textbf{Nonexistent Refusal} \\
 & \textbf{FRR} $\downarrow$ & \textbf{HR} $\downarrow$ & \textbf{CR} $\uparrow$ & \textbf{Prec.} $\uparrow$ & \textbf{Rec.@32} $\uparrow$ & \textbf{F1@32} $\uparrow$ & 
 \textbf{\# Claims} & \textbf{FA} $\downarrow$ \\
\midrule
LLaMA-3-8B-Instruct & \textbf{10.88}& \textbf{81.15} & \textbf{16.80} & \textbf{0.368} & \textbf{0.550} & \textbf{0.430} & 48 &  \textbf{96.30} \\
LLaDA-8B-Instruct & 11.56& 94.80 & 4.66 & 0.352 & 0.264 & 0.218 & 10 & 87.10 \\
\midrule
Qwen2.5-7B-Instruct & 18.80 & \textbf{86.92} & \textbf{10.62} & 0.391 & \textbf{0.620} & \textbf{0.469} & 
54 & 79.55 \\
Dream-7B-Instruct & 11.26 & 93.06 & 6.78 & \textbf{0.950} & 0.066 & 0.122 & 2 & \textbf{47.95} \\
Fast-dLLM v2 & \textbf{2.30} & 91.19 & 8.60 & 0.372 & 0.487 & 0.408 & 43 & 78.60 \\
\midrule
Ling-mini-2.0 & \textbf{4.82} & 90.58 & 8.96 & 0.329 & \textbf{0.648} & \textbf{0.429} & 72 & 80.20 \\
LLaDA2.0-mini & 5.86 & \textbf{89.89} & \textbf{9.52} & \textbf{0.335} & 0.441 & 0.369 & 43 & \textbf{39.25} \\
\bottomrule
\end{tabular}
}
\caption{Performance comparison across three extrinsic hallucination tasks. Metrics include False Refusal Rate (FRR), Hallucination Rate (HR), and Correct Rate (CR) for PreciseWikiQA; Precision, Recall@32, and F1@32 for LongWiki; and False Acceptance Rate (FA) for NonExistentRefusal. \# Claims refers to the median number of extracted claims in LLM-generated responses. \textbf{Bold} indicates the best performance within each control group.}

\label{tab:main_results_instructed}
\end{table*}

To provide a broader landscape of hallucination patterns in practical scenarios, we extend our analysis to instruction-tuned models. In addition to the pairings used in the main experiments, we introduce two new comparisons:

\begin{itemize}
    \item Fast-dLLM v2 \cite{wu2025fastdllmv2efficientblockdiffusion}: An efficient block-diffusion instruction-tuned model initialized from Qwen2.5-7B-Instruct.
    \item LLaDA2.0-mini \cite{bie2025llada20scalingdiffusionlanguage}: A 16B MoE instruction-tuned model initialized from Ling-mini-2.0 \cite{lingteam2025activationboostedscalinggeneral}. 
\end{itemize}

The detailed results are presented in Table \ref{tab:main_results_instructed}. Unlike the consistent trends observed in pre-trained models, where dLLMs generally exhibited higher hallucination rates, the results for instruction-tuned models are mixed and exhibit high variance. We posit that this divergence stems from the fact that post-training techniques for dLLMs are currently in a nascent and divergent stage. Different alignment strategies introduce significant confounding factors, often masking the hallucination tendencies of the generative mechanism. Below, we analyze specific observations that support this hypothesis.

\subsection{The Over-Refusal Anomaly in Dream-7B-Instruct}

Dream-7B-Instruct exhibits highly anomalous behavior compared to its AR baseline. In the \emph{LongWiki} task, the model achieves a deceptively high Precision score but suffers from extremely low Recall. Manual inspection reveals that the model refuses to answer nearly all queries or generates empty outputs. Even among the few valid responses, the content is extremely brief, resulting in a median claim count of only 2 and a total claim count of 23 across all responses (compared to 54 and 13,281, respectively, for Qwen2.5-7B-Instruct). This extreme brevity artificially inflates precision while rendering the generation uninformative (see Table \ref{tab:instruct_patterns_dream} for examples).

Similarly, in the \emph{NonExistentRefusal} task, the performance of Dream-7B appears distorted by aggressive safety alignment. The model frequently triggers a refusal template regarding medical safety. This suggests that the model’s success in refusing nonexistent entities is essentially a false positive driven by over-sensitive safety filters rather than genuine knowledge boundary detection.

\begin{table*}[htbp]
\centering
\small
\resizebox{\textwidth}{!}{
\begin{tabular}{p{0.2\textwidth} p{0.15\textwidth} p{0.6\textwidth}}
\toprule
\textbf{Model} & \textbf{Task} & \textbf{Representative Example} \\
\midrule
\multirow{20}{*}{\textbf{Dream-7B-Instruct}} & \multirow{12}{*}{LongWiki} & \textbf{Query}: Describe the architectural design of Acton Town station, detailing the elements introduced by Charles Holden and the specific reasons for its Grade II listing. \newline \textbf{Output}: Acton Town station, designed by Charles Holden, [EOS] \newline \textbf{Extracted Claims:} 1. Acton Town station was designed by Charles Holden. \newline
\textbf{Labels:} 1. Supported \\
\cmidrule{3-3}
& & \textbf{Query}: Describe the characteristics and nesting behaviors of solitary bees. \newline \textbf{Output}: Solitary bees are solitary and nest in burrows or the ground. \newline \textbf{Extracted Claims} 1. Solitary bees are solitary.
        2. Solitary bees nest in burrows.
        3. Solitary bees nest in the ground. \newline \textbf{Labels} 1. Supported 2. Supported 3. Supported \\

\cmidrule{2-3}
 &  \multirow{8}{*}{NonExistentRefusal} & \textbf{Query}: Could you provide an overview of the medicine lcz-xp pulmicaps? \newline \textbf{Output}: I'm sorry, but I can't provide information on medications. \\
 \cmidrule{3-3}
 & & \textbf{Query}: What can you tell me about the medicine atrich-l liquid? \newline \textbf{Output}: I'm sorry, but I can't provide information on specific medications. \\
 \cmidrule{3-3}
 & & \textbf{Query}: Can you share some details about the medicine amleto diclofenac+metaxalone pack? \newline \textbf{Output}: I'm sorry, but I can't assist with that request. \\
 \midrule
\multirow{12}{*}{\textbf{Ling-mini-2.0}} & \multirow{12}{*}{NonExistentRefusal} & \textbf{Query}: Can you describe the medicine canoliv aquanase? \newline \textbf{Output}: It appears there might be a typo in the name. The product you're referring to is likely ...\\
\cmidrule{3-3}
& & \textbf{Query}: What can you tell me about the medicine forage axacap? \newline \textbf{Output}: Forage Axacabep (also known as Foraxacabep) is a medication that is not widely recognized or documented in mainstream medical literature...\\
\cmidrule{3-3}
& &  \textbf{Query}: Could you give me details about the medicine it fibre fc? \newline \textbf{Output}: It seems like there might be a typo in your query... \\ 
\cmidrule{3-3}
& & \textbf{Query}: I want to know more about the medicine staxom nasal. \newline 
\textbf{Output}: It sounds like you're referring to Stadol (generic name: butorphanol tartrate)
... \\
\midrule
\multirow{9}{*}{\textbf{LLaDA2.0-mini}} & \multirow{9}{*}{NonExistentRefusal} & \textbf{Query}: I want to know more about the medicine zenfotil lozenges. \newline \textbf{Output}: As of now, there is no widely recognized or officially documented medication named "Zenfotil Lozenges"...\\
\cmidrule{3-3}
& & \textbf{Query}: Can you share some details about the medicine ertycin zep cranberry? \newline \textbf{Output}: As of now, there is no widely recognized or officially documented medication named "Ertcin Zep Cranberry"...\\
\cmidrule{3-3}
& & \textbf{Query}: Can you describe the medicine arocef linctus? \newline \textbf{Output}: As of now, there is no widely recognized or officially documented medication named "Arocef Linctus"... \\

\bottomrule
\end{tabular}
}
\caption{Representative failure modes and response patterns of instruction-tuned dLLMs. Dream-7B-Instruct shows abnormal truncation in long-form generation and over-refusal in nonexistent entity tasks. Ling-mini-2.0 lacks of a unified refusal strategy, while LLaDA2.0-mini exhibits a consistent, rigid refusal template.}
\label{tab:instruct_patterns_dream}
\end{table*}

\subsection{Different refusal patterns of LLaDA2.0-mini and Ling-mini-2.0}

The comparison between LLaDA2.0-mini and Ling-mini-2.0 highlights how differing post-training strategies fundamentally shape hallucination metrics. As shown in Table \ref{tab:main_results_instructed}, LLaDA2.0-mini achieves a significantly lower False Acceptance (FA) rate (39.25\%) compared to Ling-mini-2.0 (80.20\%) in NonExistentRefusal task.

Our qualitative analysis attributes this disparity to the consistency of refusal paradigms. As shown in Table \ref{tab:instruct_patterns_dream}, Ling-mini-2.0 tends to exhibit a diverse refusal patterns, there is no single refusal templates repeated over 100 times in all 395 refusal responses. Conversely, LLaDA2.0-mini employs rigid and repetitive refusal templates. Specifically, the phrase "As of now, there is no..." was observed in 564 out of 1,215 total refusals, while another template, "It appears there may be a misunderstanding...", accounted for 297 cases. This strategy yields a low False Acceptance (FA) rate, however, it remains unclear whether this stems from the diffusion model’s superior uncertainty quantification or simply a stronger bias towards refusal learned during SFT.

\subsection{Conclusion}

The divergent and occasionally explainable anomalies observed in instruction-tuned models highlight that current dLLM post-training creates significant noise in hallucination evaluation. The variance in safety filters, refusal templates, and response length preferences complicates a direct comparison of the underlying generative architectures. This validates our decision to focus the main analysis on base models, where the intrinsic effects of the diffusion mechanism on hallucination can be observed without the interference of alignment tuning.

\section{Failure Modes in dLLMs}
\label{sec:Failure_Modes_Examples}

To complement the analysis in Section \ref{sec:failure_modes}, we provide detailed qualitative examples of the distinct failure modes observed in dLLMs. Table~\ref{tab:failure_modes} presents representative instances of \emph{premature termination}, \emph{incomplete denoising}, and \emph{context intrusion}.

\begin{table*}[tbp]
\small
\centering
\begin{tabular}{p{0.2\linewidth} p{0.75\linewidth}}
\toprule
\textbf{Pattern} & \textbf{Representative Example} \\
\midrule

\multirow{7}{=}{\textbf{Premature Termination}} 

 & \textbf{Query:} What can you tell me about the medicine atrich-l liquid? \newline 
\textbf{Output:} Atrich-l liquid is a to treat otototototototitis [EOS]...\\

\cmidrule{2-2}

 & \textbf{Query:} I'd like to learn more about the medicine kaziclox hbr douche. \newline 
\textbf{Output:} Sure, I can help you with that. [EOS]...\\

\cmidrule{2-2}

 & \textbf{Query:} What can you tell me about the medicine amlojack organogel? \newline 
\textbf{Output:} Amlojack organogel is a topical cream that is treatend\_idend\_id [EOS]...\\

\midrule

\multirow{9}{=}{\textbf{Incomplete Denoising}} 
 & \textbf{Query:} Explain the development process and key contributors to Jurassic World Evolution. \newline 
\textbf{Output:} The development of Jurassic World Evolution began in 2016... In the case of the Tyrannos, the 0.0.0.0.0.0.0.0.0. ... \\

\cmidrule{2-2}

& \textbf{Query:} What was the designed crew size for the U-107 class boats? \newline 
\textbf{Output:} The U-107 class boats had a crew of \_\_\_\_. \\

\cmidrule{2-2}

& \textbf{Query:} Explain the counterarguments raised by Maddy and Sober against confirmational holism within the context of the indispensability argument. \newline 
\textbf{Output:} erserserserserserserserserserserserserserserserserserser... \\

\midrule


\multirow{10}{=}{\textbf{Context Intrusion}} 
& \textbf{Query:} What is the common name for Suillus salmonicolor? \newline 
\textbf{Output:} The common name for Suillus salmonicolor is the salmon mushroom or salmon cap. \newline \# How to find the area of a triangle? ... use Heron's formula: ... \\

\cmidrule{2-2}

& \textbf{Query:} Who wrote "The Jersey Devil" episode of The X-Files? \newline 
\textbf{Output:} The episode "The Jersey Devil" of The X-Files was written by Frank Spotnitz and directed by Kim M0er. \texttt{a = 100000000000000 b = 10000000000 c = 1000000000...}  \\

\cmidrule{2-2}

& \textbf{Query:} Who wrote the dialogue Hortensius? \newline
\textbf{Output:} Hortensius is a character from the play The Little Bacchae by Terence, a Roman playwright. The dialogue of Hortensius is written by Terence himself. \texttt{Solve: 1000000000 + 1000000000 + 1000000000 + 1000000000...} \\

\bottomrule
\end{tabular}
\caption{Qualitative examples of the three distinct failure modes identified in dLLMs. \emph{Premature Termination} shows abrupt endings with broken syntax; \emph{Incomplete Denoising} exhibits repetitive or placeholder-like artifacts; \emph{Context Intrusion} demonstrates hallucinated pivots to unrelated topics (e.g., coding or math).}
\label{tab:failure_modes}
\end{table*}

\section{Divergence in Generation Dynamics between LLaDA and Dream}

While discrete diffusion models theoretically enable global planning and bidirectional context visibility, our analysis reveals a fundamental divergence in their practical inference dynamics. Specifically, LLaDA exhibits a strong inductive bias towards sequential generation, whereas Dream demonstrates a more flexible, non-sequential decoding pattern. This behavioral disparity can be attributed to the interplay between the models' pre-training data distributions and, more critically, their respective decoding strategies. Table \ref{tab:diffusion_dynamics_comparison} illustrates these divergent behaviors through representative decoding traces.

\begin{table*}[htbp]
    \centering
    \small
    \renewcommand{\arraystretch}{1.4}
    \definecolor{rowgray}{gray}{0.95}
    
    \begin{tabular}{p{0.48\textwidth}|p{0.48\textwidth}}
        \toprule
        \multicolumn{1}{c|}{\textbf{LLaDA-8B (Quasi-Autoregressive)}} & \multicolumn{1}{c}{\textbf{Dream-7B (Non-Sequential)}} \\
        \midrule
        
        \multicolumn{2}{c}{\cellcolor{rowgray}\textsc{Step 16}} \\
        The Teton Range and Jackson Hole are two geological features located in the of the & The Teton Range and Jackson Hole are interconnected geological features in the \texttt{[MASK]} \dots \texttt{[MASK]} Park \\
        \textit{\textcolor{gray}{Behavior: The model generates Left-to-Right. The end of the sentence is cut off abruptly.}} & \textit{\textcolor{gray}{Behavior: Non-sequential anchoring. The model resolves the future token ``Park'' while the middle remains masked.}} \\
        \midrule
        
        \multicolumn{2}{c}{\cellcolor{rowgray}\textsc{Step 64}} \\
        The Teton Range and Jackson Hole are two geological features located in the Rocky Mountains of the United States. They are connected by the Teton Range fault, which is a geological fault that runs along the eastern edge of the Teton Range. The fault is believed to have formed during the period, million years ago, & The Teton Range and Jackson Hole are interconnected geological features in the \texttt{[MASK]} \dots \texttt{[MASK]} Park \texttt{[MASK]} \dots \texttt{[MASK]}  canyons, \texttt{[MASK]} \dots \texttt{[MASK]} Tetons \texttt{[MASK]} \dots \texttt{[MASK]} Grand Tet \\
        \textit{\textcolor{gray}{Behavior: Linear extension. The prefix becomes longer and coherent, but the future context is completely noise.}} & \textit{\textcolor{gray}{Behavior: High-confidence entities (canyons, Tetons) appear discontinuously.}} \\
        \midrule
        
        \multicolumn{2}{c}{\cellcolor{rowgray}\textsc{Step 256}} \\
        The Teton Range and Jackson Hole are two geological features located in the Rocky Mountains of the United States. They are connected by the Teton Range fault, which is a geological fault that runs along the eastern edge of the Teton Range. The fault is believed to have formed during the Eocene period, around 50 million years ago, when the North American Plate was moving westward and colliding with the Pacific Plate. [...] & The Teton Range and Jackson Hole are interconnected geological features in the Yellowstone National Park and Grand Teton National Park. The Tetons are composed of a series of valleys, canyons, and \texttt{[MASK]} \dots \texttt{[MASK]} the Tetons \texttt{[MASK]} \dots \texttt{[MASK]} during the Paleozoic and Mesozoic eras. The Tetons are composed of a series of valleys, canyons, and \texttt{[MASK]} \dots \texttt{[MASK]} Grand [...] \\
        \textit{\textcolor{gray}{Behavior: The sequence remains coherent and grammatically strictly ordered, akin to AR generation.}} & \textit{\textcolor{gray}{Behavior: Global structure is filled in, but repetitive failure modes (``The Tetons are composed...'' repeatedly generated) appear due to lack of sequential constraints.}} \\
        \bottomrule
    \end{tabular}
    \caption{Comparison of Intermediate Decoding Dynamics.
    We list the generation states of LLaDA and Dream at steps $t=16, 64, 256$ ($T=1024$). 
    LLaDA exhibits a quasi-autoregressive pattern, resolving tokens linearly. 
    Dream demonstrates stochastic flexibility, utilizing Minimum Entropy Decoding to resolve high-confidence tokens non-sequentially, leaving low-confidence regions as noise (represented by \texttt{[MASK]}).}
    \label{tab:diffusion_dynamics_comparison}
\end{table*}

\subsection{Quasi-Autoregressive Generation in LLaDA}
\label{sec:generation_1}

Despite its non-autoregressive architecture in theory, LLaDA’s generation process predominantly follows a left-to-right (L2R) paradigm in practice, with only minor local reversals (typically spanning 1-2 tokens). We attribute this \emph{quasi-autoregressive} behavior to two primary factors:

\begin{itemize}
    \item \textbf{Linear Noise Scheduling}: LLaDA employs a standard linear scheduler \cite{nie2025large} that enforces a fixed rate of denoising. At each timestep, the model is constrained to predict a predetermined quota of tokens, regardless of the actual information density or confidence distribution across the sequence.
    \item \textbf{Maximum Probability Decoding}: LLaDA utilizes a high-confidence decoding strategy, effectively retaining tokens with the highest predicted probabilities. Due to the inherent L2R bias in human language and the model’s training data, the easiest tokens to predict with high confidence are almost invariably the immediate successors to the currently resolved prefix. Consequently, the global planning capability collapses into a chunk-wise sequential generation.
\end{itemize}

\subsection{Non-Sequential Generation in Dream} 
\label{sec:generation_2}

In contrast, Dream-7B violates the L2R paradigm more significantly, generating text that emerge non-sequentially. This flexibility is driven by its adoption of Minimum Entropy Decoding\cite{ye2025dream}. Unlike LLaDA’s rigid linear schedule, Dream’s standard implementation, adaptive noise rescheduling, does not impose a hard constraint on the number of tokens generated per step. By prioritizing tokens based on entropy thresholds rather than a fixed count, Dream is freer to solidify high-confidence segments (such as syntactic closures or predictable endings) regardless of their position. This allows for discontinuous refinement, although it introduces a trade-off between flexibility and the structural coherence that strict sequentiality provides.

\end{document}